\documentclass[10pt,twocolumn]{article} 

\usepackage{simpleConference}
\usepackage{times}
\usepackage{graphicx}
\usepackage{amssymb}
\usepackage{url,hyperref}
\usepackage{threeparttable}
\usepackage{multirow}
\usepackage[namelimits]{amsmath}

\begin{document}

\title{ULDGNN: A Fragmented UI Layer Detector Based on Graph Neural Networks}
    \author{
        Jiazhi Li\\
		Zhejiang University\\
		Hangzhou, China\\
		{\tt\small lijz@zju.edu.cn}
		\and
		Tingting Zhou\\
		Alibaba Group\\
		Hangzhou, China\\
		{\tt\small miaojing@taobao.com}
		\and
		Yunnong Chen\\
	    Zhejiang University\\
		Hangzhou, China\\
		{\tt\small chen\_yn@zju.edu.cn}
		\and
		Yanfang Chang\\
		Alibaba Group\\
		Hangzhou, China\\
		{\tt\small suchuan.cyf@alibaba\-inc.com}
		\and
	    Yankun Zhen\\
		Alibaba Group\\
		Hangzhou, China\\
		{\tt\small     zhenyankun.zyk@alibaba\-inc.com}
		\and
		Lingyun Sun\\
		Zhejiang University\\
		Hangzhou, China\\
		{\tt\small sunly@zju.edu.cn}
		\and
		Liuqing Chen\\
		Zhejiang University\\
		Hangzhou, China\\
		{\tt\small chenlq@zju.edu.cn}
	}

\maketitle

\begin{abstract}
 While some work attempt to generate front-end code intelligently from UI screenshots, it may be more convenient to utilize UI design drafts in Sketch which is a popular UI design software, because we can access multimodal UI information directly such as layers type, position, size, and visual images. However, fragmented layers could degrade the code quality without being merged into a whole part if all of them are involved in the code generation. In this paper, we propose a pipeline to merge fragmented layers automatically. We first construct a graph representation for the layer tree of a UI draft and detect all fragmented layers based on the visual features and graph neural networks. Then a rule-based algorithm is designed to merge fragmented layers. Through experiments on a newly constructed dataset, our approach can retrieve most fragmented layers in UI design drafts, and achieve 87\% accuracy in the detection task, and the post-processing algorithm is developed to cluster associative layers under simple and general circumstances.
\end{abstract}

\section{Introduction}
Graphic User Interface (GUI) builds a visual bridge between software and end users through which they can interact with each other. A good GUI design makes software more efficient and easy to use, which has a significant influence on the success of applications and the loyalty of its users. \par
\begin{figure}[!htb]
\centering 
\includegraphics[scale=0.37]{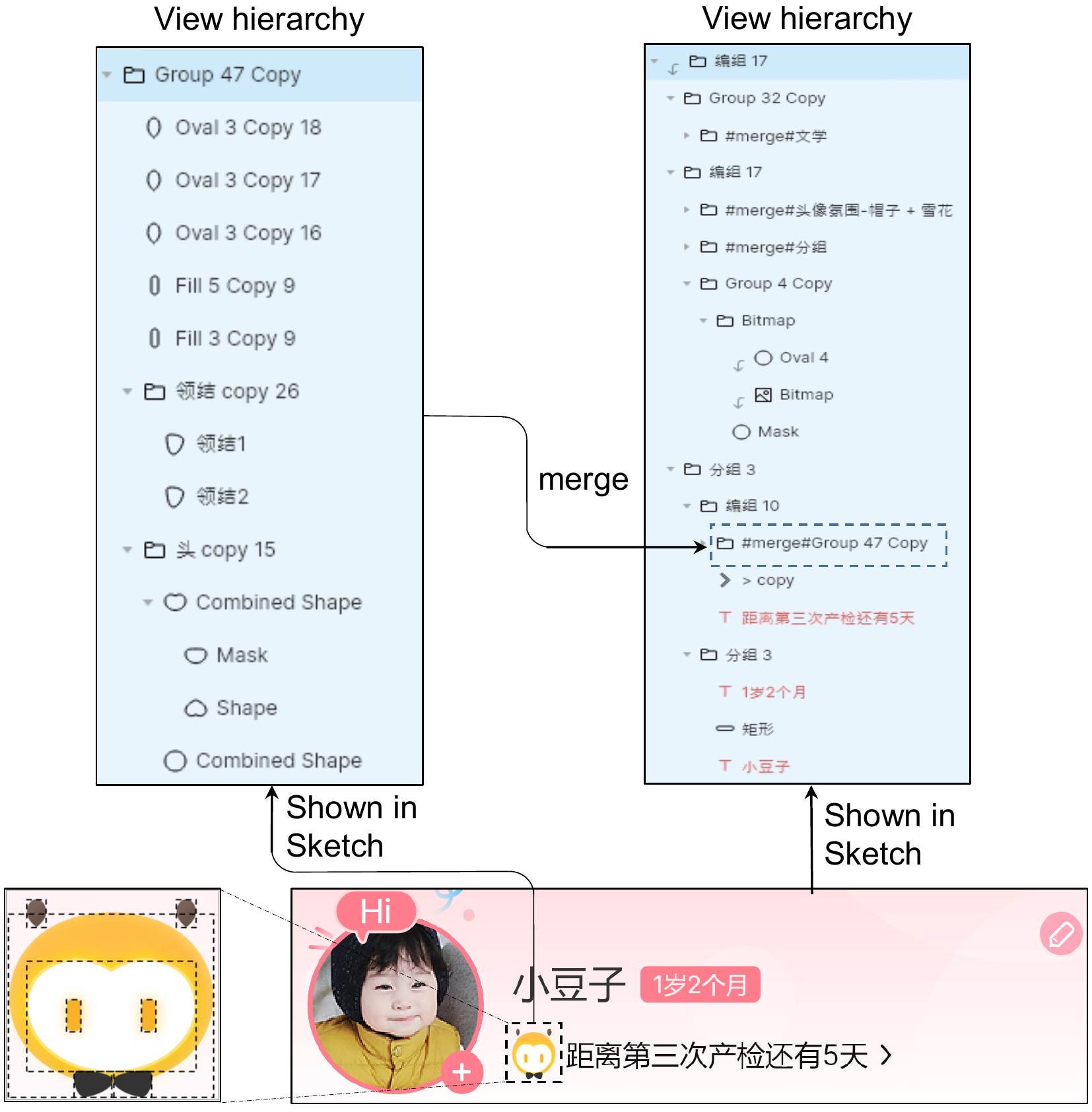} 
\caption{The original view hierarchy is complicated before merging fragmented layers in the UI icon and it is simplified a lot after merging layers and using a single node to represent a UI icon. The dashed rectangles in the UI icon represent fragmented layers.}
\label{layoutmergingresult} 
\end{figure}

This manuscript aims to detect and correct issues in the GUI design drafts automatically, which is the first step of intelligent front-end code generation. Some previous work has already attempted to adopt the deep learning technique to generate maintainable UI view code and logic code automatically from design drafts \cite{pix2code,GUIfetch,imgcook}. The main reason for intelligent code generation is that GUI implementation is a time-consuming process for developers and they can hardly devote majority of the time to developing unique features of an application or website. To free the front-end developers from tedious and repetitive work, researchers aim at developing an intelligent code generation system in which the quality of the original design draft plays a vital role. However, the designers mainly focus on the aesthetic of their design and usually ignore some design standards. It is important to correct them before the intelligent code generation. \par
We are mainly concerned about correcting UI layer merging issues in design drafts which contain rich multimodal UI information and are usually produced by designers with UI design software such as Sketch. We can directly access each layer's type, position, size, and visual images by traversing UI in a design draft. 
The main reason of UI merging issues is that UI designers usually stack layers to design their creative patterns such as
a background pattern and a UI icon, 
which not only increases the burden of UI layout calculation but also impairs the maintainability of generated code. The whole pattern should be implemented as a single element. For example, as shown in Fig.\ref{layoutmergingresult}, the view hierarchy in a design draft is very complicated without merging fragmented layers in the UI icon. After the merging process, the UI layout is simplified greatly as we can use a single node to represent a complicated pattern. Furthermore, as shown in Fig.\ref{codegeneration}, the generated DOM tree without merging fragmented layers has complex nested structure and it is simplified a lot after merging layers in the UI icon, which improves the readability and maintainability of generated front-end code a lot. In summary, it is significant to merge fragmented layers in design drafts to guarantee generating front-end code of high quality. However, it is a great challenge to detect and merge associative fragmented layers.
\par
As described above, we aim to propose an approach to detect and merge all associative fragmented layers in a design draft, which can improve the maintainability and readability of corresponding generated code. We try to solve the problem in two steps. Firstly, all layers in the UI design draft are classified into two classes by a deep learning method, which can judge whether a layer is fragmented or not. We then design a rule-based post-processing algorithm to cluster the predicted trivial layers which belong to the same pattern.\par

Specifically, we propose a model composed of two main parts for detecting fragmented layers, which are a backbone network and a graph neural network model. The backbone network is used to encode multimodal information of local regions within each layer’s boundary. Each design draft has an underlying hierarchy to organize layers. The hierarchy is represented by a JSON file. We first traverse the hierarchy and construct a new graph structure for it based on each layer's position and size information. Each node in the graph represents a layer in the UI layout and its state vector is initialized with the multimodal information encoded by the backbone network. The graph neural networks update and refine vertex states through a message-passing framework. In this way, every layer finally knows about the neighbors around it and how it belongs to the UI layout. We construct a graph for the UI layout and use a GNN model to learn each layer's features because layers in the same pattern usually have strong relationships with each other. They may have similar shapes, colors, and geometrical relations such as parallelism and inclusion. To fully exploit the relation between layers, we train the GNN model so that each layer can obtain a better final representation that contains information about how it interacts with others. An MLP classifier is finally utilized to classify each layer based on the learned representation vectors.\par

We conduct experiments on a proposed dataset containing 4644 graphs to evaluate our proposed model and explore the performance of different CNN backbones and GNN models. We also visualize the results of our post-processing algorithm where layers in the same pattern are bounded with the same color. Through quantitative and qualitative analysis, our work can serve as a basic approach to solving the layer merging problem. In summary, the main contributions of this paper are:\par
\begin{itemize}
    \item This is the first work trying to detect and merge fragmented layers in design drafts. We also construct an open dataset which is used for the evaluation of our proposed method.
    \item We propose a method of constructing the graph for a UI design draft and utilizing a GNN model to solve the layer merging problem based on both visual information and underlying multimodal hierarchy information.
    \item We propose a UI layer merging pipeline to cluster fragmented layers contained in the same pattern, which can simplify and facilitate the high-quality UI code generation process.
\end{itemize}
\label{sec:introduction}

\section{Related work}
\subsection{Intelligent Code Generation}
It is time-consuming and tedious to implement GUI code. To facilitate GUI code generation, researchers turn to machine learning techniques trying to generate code intelligently. Nguyen et al. \cite{ReverseEngineering} identified UI elements in a given input bitmap via computer vision and OCR technique to reverse-engineer Android user interfaces. Beltramelli et al. \cite{pix2code} and Jain et al. \cite{jain2019sketch2code} generated computer tokens from GUI screenshots based on Convolutional and Recurrent Neural networks. Chen et al. \cite{UI2skeleton} proposed a neural machine translator to translate a UI design image into a GUI skeleton which would be beneficial for bootstrapping mobile GUI implementation. Moran et al. \cite{ML-Prototyping} presented a data-driven approach for prototyping software GUIs automatically which was capable of detecting GUI components and hierarchy generation. Chen et al. \cite{storyboard} automatically generated a visualized storyboard of Android apps by extracting relatively complete ATG. Zhao et al. \cite{GUIGAN} utilized a generative adversarial network to generate GUI designs automatically. Some work \cite{GUIfetch,SeekUI} helps develop an app UI more quickly by a code retrieval-based approach.

\subsection{UI Issue Detection}
To guarantee the quality and correctness of generated code, there are some code static linting tools to correct programming errors and style errors. For example, styleLint \cite{stylelint} is an open-source plugin which helps detect errors and enforces style conventions based on over 170 built-in rules. Some work \cite{owleye} detects the final GUI displaying issues, such as text overlaps, blurred screens, and missing images, to guide developers to fix the bug. 
Chen et al. \cite{unblindyourapp} developed a deep learning based model to predict missing labels of image-based buttons. Zhao et al. \cite{seenomaly} formulated the GUI animation linting problem as a classification task and proposed an auto-encoder to solve the linting problem. Different from static linting, Baek et al. \cite{AutomatedModelBased}, Mirzaei et al. \cite{ReducingCombination} and Su et al. \cite{GuidedStochasticMBGUItest} can dynamically analyze a mobile app GUI by a multi-level GUI Comparison Criteria. 
Several surveys \cite{lamsa2017comparison,systematicMappingStudy} compared different tools used for GUI testing of mobile applications. Recently, Degott et al. \cite{degott2019learningUIelement} adopted the reinforcement learning technique for automatic GUI testing. White et al. \cite{white2019improvingRandomGUI} utilized computer vision techniques for identifying GUI widgets in screenshots to improve GUI test.
\begin{figure*}[]
\centering 
\includegraphics[scale=0.55]{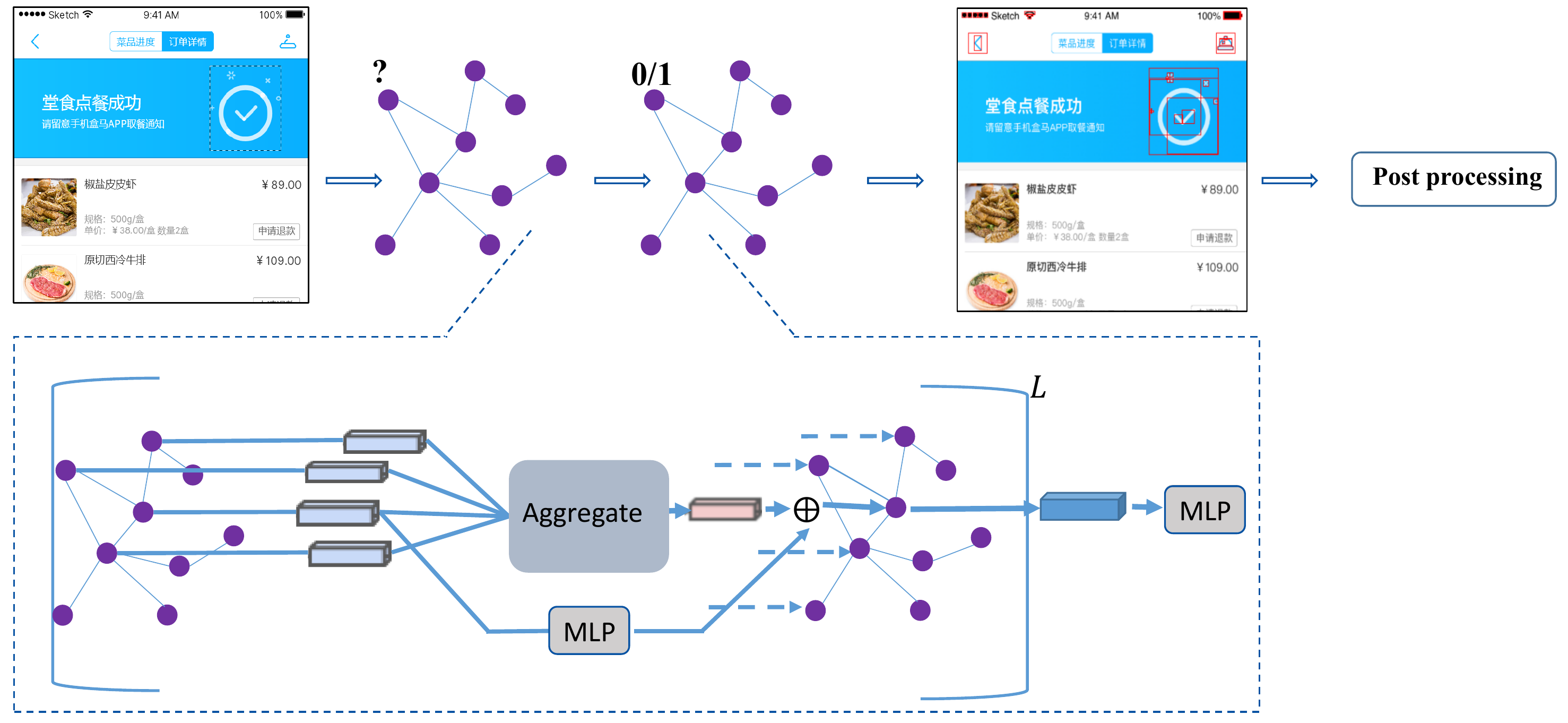} 
\caption{\textbf{Architecture overview}: the UI layout is converted into a graph and each node represents a layer. Then a graph neural network is utilized to classify each node after which a post-processing algorithm is designed to merge associative layers.} 
\label{fig5} 
\end{figure*}

\subsection{Graph Neural Networks}
Recent years have witnessed a great surge of promising graph neural networks (GNN) being developed for a variety of domains including chemistry, physics, social sciences, knowledge graphs, recommendations, and neuroscience. The first GNN model was proposed in \cite{gori2005new}, which is a trainable recurrent message passing process. To generalize the convolution operation to non-Euclidean graphs, these work \cite{bruna2013spectral,defferrard2016convolutionalneuralNetworkongraphs,kipf2016GCN} defined spectral filters based on the graph laplacian matrix. The learned filters in these spectral approaches depend on the graph structure, so they cannot generalize to a graph with different structures. Spatial-based models define convolutions directly on the graph vertexes and their neighbors. Monti et al. \cite{monti2017geometric} presented a unified generalization of CNN architectures to graphs. Hamilton et al. \cite{GraphSAGE} introduced GraphSAGE, a method for computing node representations in an inductive manner which operates by sampling a fixed-size neighborhood of each node and performing a specific aggregator over it. Some methods attempted to enhance the original models with anisotropic operations on graphs, such as attention \cite{velivckovic2017gat} and gating mechanisms \cite{bresson2017residualgatedGCN}. Xu et al. \cite{xu2018GIN} aimed at improving upon the theoretical limitations of the previous model. Li et al. \cite{li2019deepgcns} and Chen et al. \cite{chen2020GCNII} tried to overcome the over-smoothing problem when GCN goes deeper. \par
Researchers also have great interest in utilizing graph neural networks to tackle computer vision tasks, such as 3d object detection \cite{shi2020pointGNN}, skeleton-based action recognition \cite{wen2019graphskeleton}, semantic segmentation \cite{qi20173d}. Inspired by these work, we also attempt to introduce a graph neural network model to our proposed pipeline to detect issues in the UI design drafts. More details are described in Section 3. 
\label{sec:related}

\section{Methodology}
In this section, we describe the proposed approach exploiting the rich multimodal information to detect the UI issues described in Section 1. As illustrated in Fig.\ref{fig5}, we first construct a graph representation for the UI layout in which each node stands for a layer (Section 3.2). All nodes are initialized with their encoded multimodal information as described in Section 3.3. We detect all layers which should be merged through a graph neural network followed by an MLP classifier (described in Section 3.4). It is a two-class classification task. A post-processing algorithm described in Section 5.1 is then designed to cluster the positive samples. After fragmented layers are detected and clustered, which should be merged, they are combined into UI components within some specific context and can be represented by a single layer which can significantly reduce the number of layers in the UI design drafts and help improve the quality of generated front-end code.
\subsection{Data Introduction}
The input data to our algorithm are artboards contained in files with .sketch suffix. The files are created by Sketch which is a popular UI draft design software. Each artboard represents the UI design of the Android or iOS mobile phones. The layers in the artboard are organized by a tree structure. We traverse the tree in a pre-order manner to get all layers which are represented by leaf nodes. 
Each layer has its own multimodal information including type, size, position, and visual images. The size and position are utilized to construct a graph representation for the UI layout and all multimodal information is encoded into feature vectors to initialize each node in the graph. 
\begin{figure*}[!htb]
\centering 
\includegraphics[scale=0.4]{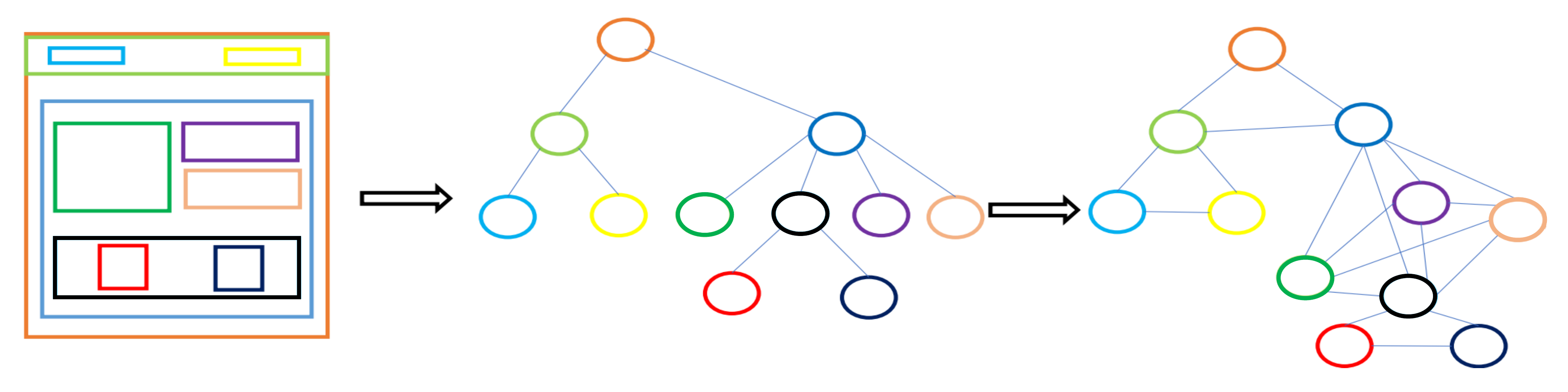} 
\caption{\textbf{Graph construction}: The UI layout is first converted into a tree and we connect edges between sibling nodes. Each node represents a layer and the tree is constructed based on the inclusion relationship.} 
\label{graphconstruction} 
\end{figure*}
\subsection{Graph Construction}
After the traversal of the artboard, we obtain all layers’ type, position, and size information. We construct a new tree based on the inclusion relationship between layers. The tree can be constructed online while traversing the artboard. In the beginning, an empty root node is constructed which denotes an empty canvas. When layer A is visited, it is inserted into the tree and it will finally be added into node B’s children if node B’s bounding box contains A while none of B’s children contains A. \par
As described in Fig.\ref{graphconstruction}, a directed graph is constructed based on the new tree representing the UI layout. Actually, a tree data structure is a special kind of graph, so we keep the directed edges in the tree and meanwhile, we construct a complete undirected graph for those sibling nodes in the tree. We use the inclusion relationship to construct a tree due to the following considerations: 1) layers combined into a background usually are contained by a large layer, 2) an empty layer contains other layers to represent a UI component, 3) UI designers usually use some empty layers to split the UI layout into different parts and layers in different parts will never be merged together. The generated graph based on the new tree cuts off interactions between layers in different parts. However, there exist some circumstances, for example, a WIFI icon, under which layers are ranged in a straight line or spread over the entire UI layout and they are not contained by some large layers but can be merged into one component within some specific context. Therefore we just construct a complete graph for those sibling nodes in the tree so that the GNN model can detect these components automatically.
\subsection{Feature Extraction}
\begin{figure}[h]
\centering 
\includegraphics[scale=0.32]{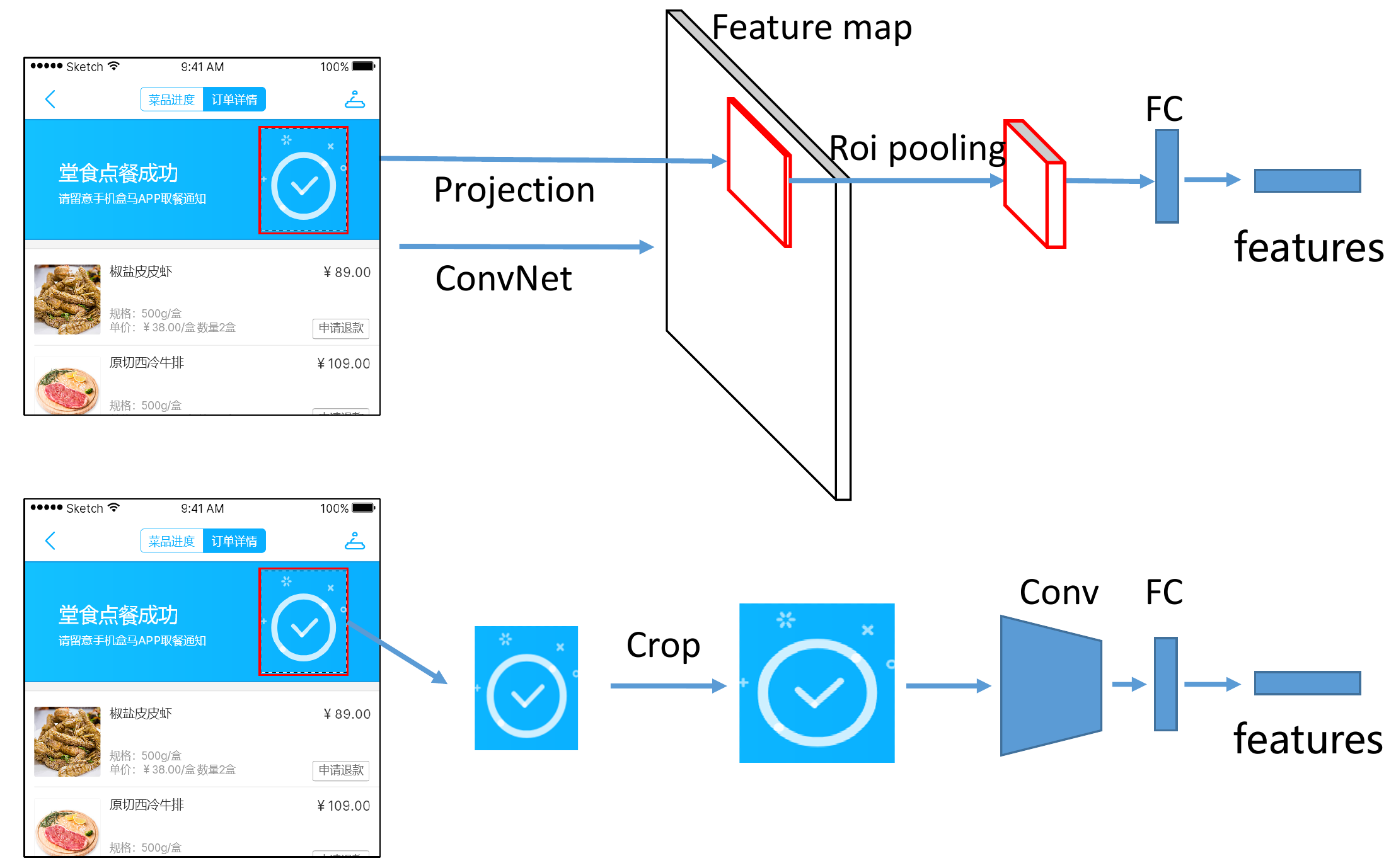} 
\caption{The first row shows that we extract visual features from corresponding regions in the feature map with RoI pooling. The second row shows that we propagate the cropped layer image through a CNN backbone to obtain its visual features.} 
\label{feat_extraction} 
\end{figure}
In this section, we describe how to encode the multimodal information to initialize each node in the graph. \par
We use a one-hot code to denote the layer's type and embed it into the embedding space by a parameter matrix. For the position and size of layers, we just simply use another matrix to encode the information. All vectors are concatenated together to form an inherent representation of each layer. \par
It is hard to classify layers only based on the type, position, and size information. We human beings judge the UI issues mainly from shapes, colors, or some other visual features of layers in the UI layout. And we can use vision knowledge to infer the context which helps us detect issues. It is even a challenge for human beings to detect which layers should be merged together based only on the type, size, and position information. Thus it is critical to extract each layer’s rich visual information from the UI screenshots. \par
There are two methods to extract corresponding features from the images based on the layer’s position and size. As shown in Fig.\ref{feat_extraction}, the first straightforward way is that we transform the image data inside the layer region to a fixed size regardless of the aspect ratio, and the visual features are computed by propagating the RGB matrix through a convolutional neural network followed by one fully connected layer. Inspired by \cite{girshick2015fastrcnn}, the second method is that we can use the RoI pooling operation to extract features. Firstly, the UI screenshot is fed into a CNN backbone to obtain a feature map. We then divide the corresponding layer window with size $H \times W$ into $h \times w$ grids. Each grid with size $H/h \times W/w$ is max pooled to output a maximum value. In this way, we can convert the features inside any valid layer region into a small feature map with a fixed spatial extent of $h \times w$. Both $h$ and $w$ are set to be 5 in our experiments. A fully connected layer is finally adopted to obtain a high-dimensional feature vector for each UI layer. \par
The three feature vectors described above are concatenated together to form the initial representation of the nodes in the graph. Through the graph neural network model, each node in the graph finally learns how it belongs to the graph based on the types, positions and visual features of its neighbors.\par
\begin{table*}[!ht]
\caption{Comparisons of different CNN and GNN combinations on four evaluation metrics}
    \centering
    \setlength{\tabcolsep}{7mm}{
    \begin{tabular}{l|c|c|c|c}
    \hline
        method & precision & recall & accuracy  & f1-score \\ \hline\hline
        VGG16+GAT & \textbf{0.880} & 0.868 & 0.867 & 0.874 \\ 
        VGG16+GIN & 0.862 & 0.860 & 0.852 & 0.861 \\ 
        VGG16+GCN & 0.855 & \textbf{0.886} & 0.860 & 0.871 \\ 
        VGG16+GCNII & 0.856 & 0.875 & 0.860 & 0.865 \\
        ResNet50+GAT & 0.860& 0.876 & 0.862 &  0.868 \\ 
        ResNet50+GIN & 0.860 & 0.879 & 0.863 & 0.869 \\ 
        ResNet50+GCN & 0.862 & 0.873 & 0.864 & 0.869 \\ 
        ResNet50+GCNII & 0.867 & 0.884 & \textbf{0.870} & \textbf{0.875} \\ \hline
    \end{tabular}
    }
    
    \label{experiment}
\end{table*}
\subsection{Graph Neural Networks}
A typical graph neural network can be implemented by a messaging-passing framework. Every node updates its state mainly through three steps in the form:
\begin{equation}
\begin{aligned}
&\textbf{m}_{ij}=MESSAGE(\textbf{h}_i,\textbf{h}_j)\\
&a_i = AGGREGATE({\textbf{m}_{ij}:j\in\mathcal{N}_i})\\
&\textbf{h}_i^{'}=UPDATE(\textbf{h}_i,a_i)
\end{aligned}
\end{equation}
Each neighbor of the node prepares a message with \textit{MESSAGE} function to propagate along the edge. Then each node uses an \textit{AGGREGATE} function to integrate all messages from its neighbors, and updates its state based on the messages received and the state itself. Usually \textit{MESSAGE} and\textit{ UPDATE }function are implemented as a multi-perceptron network and the \textit{AGGREGATE} function should be a permutation invariant function to eliminate the influence of message input order.\par
Each neighbor is involved in the state updating process differently because layers in the same UI component have more influence on each other than layers outside. Considering the various importance of layers, we introduce the attention mechanism in \cite{velivckovic2017gat} to our model to learn the importance of different neighbors automatically. It uses a weight matrix $ \textbf{W} \in {R}^{F\times {F}^{'}}$ to embed all nodes’ feature vectors through a linear mapping. A shared parameter $ \textbf {a} \in {R}^{2F^{'}} $ is adopted to calculate the attention coefficients in the form:
\begin{equation}
e_{ij}=\sigma(a(\textbf{W}\textbf{h}_i,\textbf{W}\textbf{h}_j))
\end{equation}
$\sigma(.)$ is a LeakyReLU activation function. The coefficients are normalized by a softmax function:
\begin{equation}
\alpha_{ij}=softmax_j(e_{ij})=\frac{exp(e_{ij})}{\sum_{k\in\mathcal{N}_i}exp(e_{ik})}
\end{equation}
$\alpha_{ij}$ can be used to measure how important of node i is  to node j. The update function of node i can be written as:\par
\begin{equation}
\textbf{h}_i=\sigma(\sum_{j\in\mathcal{N}_i}\alpha_{ij}\textbf{W}\textbf{h}_j)
\end{equation}\par
Another reason for using graph attention layers in our model is that we need to deal with graphs with diverse topologies. Our model should generalize well to design drafts which are unseen during the training phase because completely different graphs are generated for different UI layouts. For the inductive learning task, it is not suitable to use spectral-based graph neural networks which cannot generalize to graphs with unseen topology. Graph attention networks, however, can directly be applied to inductive learning and can deal with directed graphs. We replace the graph attention model with different spatial-based models to observe their performance in this task. \par
Through the graph neural network, the final state vectors of each node are fed into an MLP classifier to judge whether a layer should be merged or not. 
\label{sec:methodology}

\begin{figure}[]
\centering 
\includegraphics[scale=0.5]{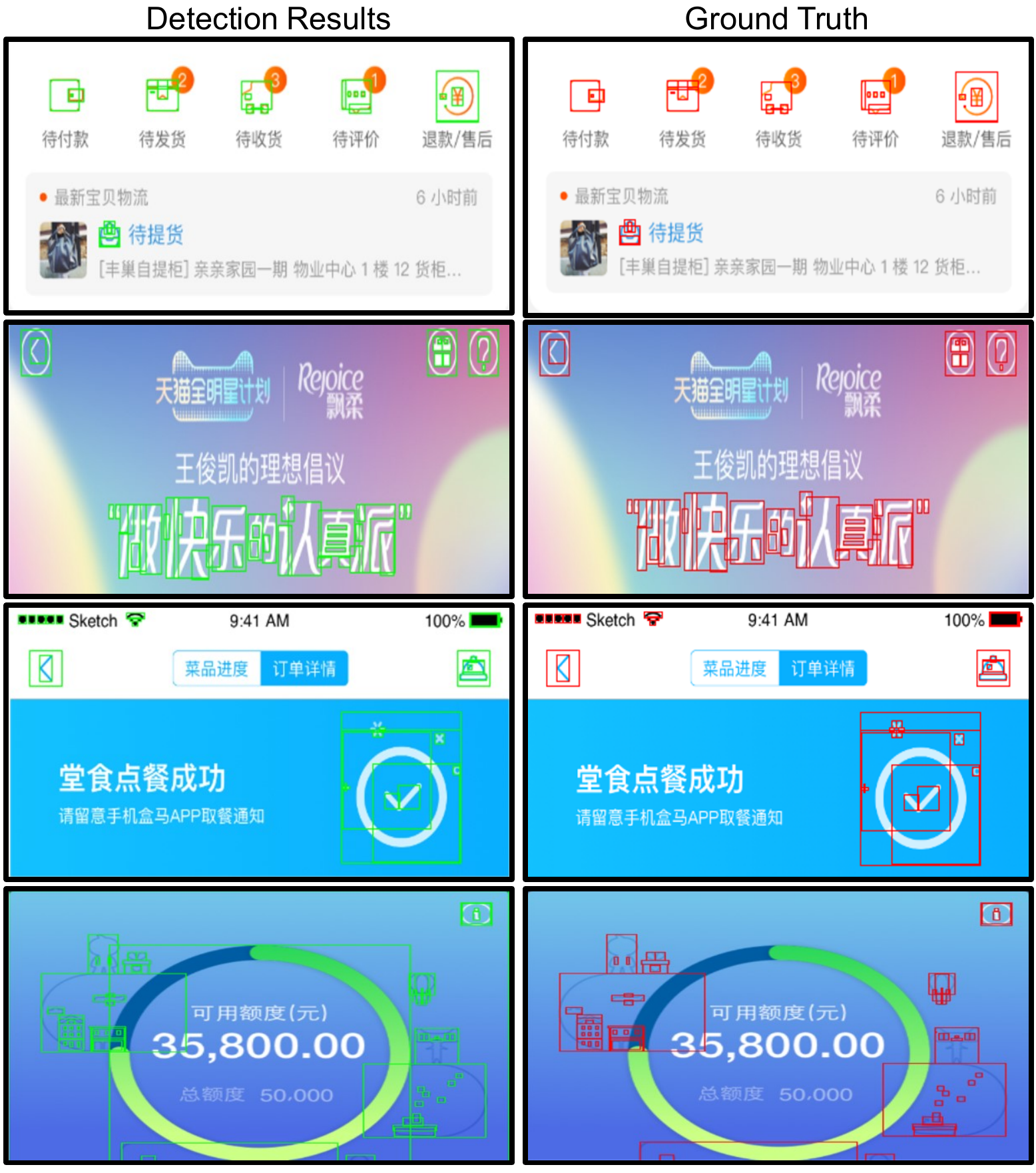} 
\caption{\textbf{Detection results}: each fragmented layer are bounded by a rectangle. Fragmented layers may be merged to form a UI icon, a decorative pattern or art font, and a background pattern.}
\label{qualitativeresult} 
\end{figure}

\section{Experiments}
\subsection{Implementation Details}
\textbf{Data Generation}: GUI in mobile Apps are usually scrolled vertically and layer relations are often within local regions. Moreover, a complete artboard may consist of many trivial layers. If we construct a graph for a complete artboard based on the algorithm described in Section 3.2, the large-scale graph may not even fit into the GPU memory. Layers rarely have a long-distance dependency on each other, that is, a layer has a weak relationship with those which are far from it. Given the reasons above, we can split an artboard into small patches and the model can look into the local regions to detect fragmented layers. Specifically, we scale the height and width of the artboard to a multiple of 750. The artboard is then divided into small windows of size 750×750. A layer will be included in a window if its center is located inside the window. The graph construction algorithm converts the UI layout in the window into a graph which is then fed into the GNN model. Each artboard has a corresponding image which is resized and cut into small patches in the same way. The images serve for each layer’s visual feature extraction. \par
As the graphs generated in the same artboard may have great similarity, we split the dataset in granularity of artboard to test the model’s generalization ability and we use 80\% of the dataset as the training set, 10\% as the validation set, and the last 10\% as the test set. Finally, a total of 4644 graphs are used for training, and 1048 graphs for validation and test respectively.\par
\noindent\textbf{Details of Training}: We adopt the VGG16 model pre-trained on the ImageNet dataset as our visual feature extraction backbone. We also conduct experiments by replacing VGG16 with ResNet50. The convolutional neural networks are also trained to adapt to our task. The first ten layers of VGG16 and the first three residual blocks of ResNet50 are fixed during training. The graph attention networks (GAT) consist of four graph attention layers. To stabilize the training process, we use multi-head attention in each layer similar to \cite{velivckovic2017gat}. In the first three layers, each has 4 attention heads and we concatenate them into a single vector which is then followed by ELU non-linearity. For the last layer, it has 6 attention heads and we average the output. We add a skip connection from the previous attention layer to the current one. The skip connection structure consists of a single MLP layer. We utilize the Adam optimizer with an initial learning rate of 1e-3 and it is reduced by half if the validation loss does not change for 10 epochs until it reaches the minimum value of 1e-6. We implement our algorithm with the Pytorch and Pytorch-Geometric library. \par
\noindent\textbf{Evaluation Metric:} We evaluate and compare different models' performance for UI issue detection with four commonly used metrics: accuracy, precision, recall, and f1-score. \par 
\subsection{Results}
\textbf{Quantitative Analysis:} Our GNN model is used to detect fragmented layers in the UI layout which should be merged with others. It is an inductive learning task in which GNN models generalize well to unseen graphs. Therefore GNN models are implemented by a spatial-based method which usually is based on a message-passing framework. We conduct experiments on different GNN models and report their performance based on various evaluation metrics. We use the first visual feature extraction method introduced in Section 3.3. For popular GCN in \cite{kipf2016GCN}, it consists of three graph convolution layers which are implemented in a spatial-based manner. For fairness, we also add the same skip connection to the GCN model. For the GCNII model introduced in  \cite{chen2020GCNII} and the GIN model \cite{xu2018GIN} we use the same configuration in the original papers. To combine different advantages of CNN and GNN, we concatenate the output of CNN and GNN to form the final representation vector used for classification. As shown in Table \ref{experiment}, with VGG16 as the backbone network, the GAT model performs best on most evaluation metrics. The GCN model and GCNII model can outperform other GNN models regarding the recall metric. Although the GIN model is as powerful as WL-test, it can not well adapt to our fragmented layer detection task when combining VGG16 model. When replacing VGG16 with ResNet50 as the backbone network, most GNN models have a slight increase in recall and accuracy metrics. Although the combination of ResNet50 and GCNII outperforms all the other combinations on accuracy and f1-score, it just has a little improvement on the fragmented layer detection task. We finally adopt the combination of VGG16 model and GAT model because they have fewer parameters and training time while achieving comparable performance. It seems that different GNN models perform similarly on the four metrics, however, they play a vital role in the fragmented layer detection task. We compare the performance of a single CNN and a combination of CNN and GNN to validate the effectiveness of the GNN models. More details will be described in Section 4.3.   \par
\begin{figure}[]
\centering 
\includegraphics[scale=0.4]{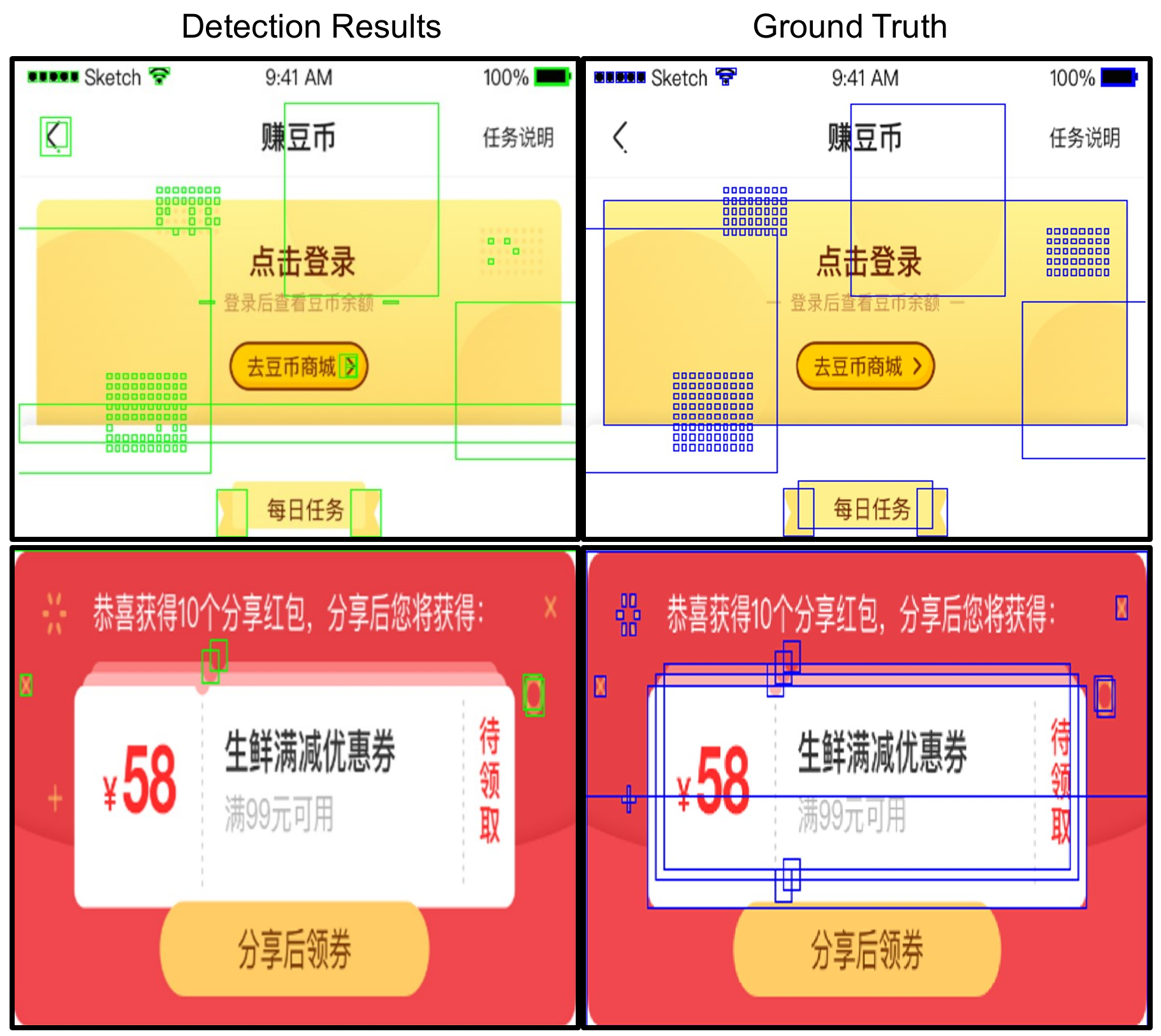} 
\caption{The are some cases in which our model is difficult to detect all fragmented layers. }
\label{failurecases} 
\end{figure}
\noindent\textbf{Qualitative Analysis:} We visualize some detection results in the test dataset as shown in Fig.\ref{qualitativeresult}. Each fragmented layer, which should be merged, is marked with a green rectangle. We can see that designers usually stack layers to draw a UI icon, a decorative pattern or art font, and a background pattern.
These fragmented layers should be merged into a whole part before generating front-end code intelligently. The merging process, which decreases the number of layers a lot, not only eases the burden of layout calculation in the intelligent code generation but improves the quality and maintainability of generated front-end code. Our model can detect a majority of fragmented layers in UI icons, decorative patterns, art fonts and complicated background patterns, and it can serve as a useful visualization tool to remind the designers of merging fragmented layers manually. In Section 5, we design and evaluate a simple post-processing algorithm to merge trivial layers detected by our model to avoid repetitive and time-consuming human effort for merging layers. 

\noindent\textbf{Failure Cases Analysis: } While our approach can detect the majority of trivial layers without specific context in a UI icon or a simple decorative pattern, there are still some challenging cases where our algorithm performs not well as shown in Fig.\ref{failurecases}. For some complicated background patterns which are composed of many tiny layers, it is hard to retrieve all fragmented layers as the layers are spread over the entire background. One possible explanation is that it requires some cognitive knowledge about UI aesthetics and UI design to correctly classify and merge fragmented layers in the background. For example, in Figure 6, human beings conclude that layers bounded by blue rectangles are fragmented mainly based on some high-level knowledge, that is UI visual aesthetic. However, it is difficult for machines to learn about such knowledge, which is a great challenge for classifying layers correctly in the background. In Fig.\ref{failurecases} we also notice that our model fails to detect the large background layer because the layer is occluded by foreground layers. Furthermore, layers may be combined by a boolean operation to form some specific shapes but it does not matter that our model fails to detect these layers, because we can merge them directly if we know layers are combined by a boolean operation. 
\subsection{Ablation Study}
\begin{table}[!ht]
    \caption{The performance of our pipeline with and without GNN}
    \label{ablation1}
    \centering
    \setlength{\tabcolsep}{1mm}{
    \begin{tabular}{c|c|c|c|c}
    \hline
        method & precision & recall & accuracy  & f1-score \\ \hline
        VGG & 0.844 & 0.819 & 0.829 & 0.831\\
        VGG+GAT & \textbf{0.880} & \textbf{0.868} & \textbf{0.867} & \textbf{0.874} \\
        \hline
    \end{tabular}}
\end{table}
\noindent\textbf{Single CNN vs CNN+GNN:} We remove the GNN model while keep the MLP classifier to validate the effectiveness of GNN. Table \ref{ablation1} reports the performance of the single CNN with the first visual extraction method. We can see that the combination of GNN and CNN outperforms the single CNN model by 4\% regarding the accuracy. And with GNN model, we can retrieve more fragmented layers as the table shows a great increase of recall by about 5\%. Due to the great improvement on the four metrics, we can conclude that the GNN model plays a vital role in the detection task. Fragmented layer prediction with a single CNN model only focus on features within local regions which ignores layer's relations with others. It tends to learn the common features of the fragmented layers. When using the spatial-based graph neural networks, each node in the graph can update its own state according to the features of its neighbors. Each layer in the UI layout can finally learn how to interact with layers around, which can benefit for detecting fragmented layers a lot. \par
\begin{table}[!ht]
    \caption{Comparisons between two visual feature extraction methods.}
     \label{ablation2}
    
    \centering
     \setlength{\tabcolsep}{1.2mm}{
      \begin{threeparttable}
     \begin{tabular}{c|c|c|c|c}
     \hline
        method & precision & recall & accuracy  & f1-score \\ \hline
        RoI\tnote{1} & 0.835 & 0.835 & 0.825 & 0.835 \\ 
        CropCNN\tnote{2} & \textbf{0.880 }& \textbf{0.868} & \textbf{0.867} & \textbf{0.874} \\ 
        \hline
    \end{tabular}
    \begin{tablenotes}
    \footnotesize
    \item[1] RoI means that the method extracts local features by the RoI pooling operation. 
     \item[2] CropCNN means that we use CNN to extract features from cropped layer image directly.
     \end{tablenotes}
    
    \end{threeparttable}
    }

\end{table}
\noindent\textbf{Visual Feature Extraction:} We compare the performance of two visual feature extraction methods described in Section 3.3. The results in Table \ref{ablation2} show that using the CNN backbone to extract features directly from the corresponding local region performs better in our task, which outperforms the method of using the RoI pooling operation greatly. One possible explanation is that the layers which belong to the same part usually have some similar visual features such as color and shapes. Layers with similar visual features should be represented by feature vectors with great similarity. The key difference between the two feature extraction methods is where the extracted features come from. For the RoI pooling method, it extracts features from the feature map obtained by propagating the image through a CNN backbone. Similar layers in different positions obtain different feature representations because different local regions in the feature map have great dissimilarity, which explains the poor performance of the RoI pooling method. As for the method of using a CNN backbone directly, it extracts features from the original image directly. Layers with similar visual features obtain similar representations because we propagate similar cropped layer images through the same CNN backbone. Therefore, the nodes in the graph can be better initialized and our model detects fragmented layers better when adopting the method of extracting features from original images. 

\par
\begin{table}[!ht]
    \caption{Comparisons between different feature fusion strategies.}
    \label{ablation3}
    \begin{threeparttable}
    \centering
    \begin{tabular}{c|c|c|c|c}
    \hline
        method & precision & recall & accuracy  & f1-score \\ \hline
        LE\tnote{1} & 0.818 & \textbf{0.910} & 0.85 & 0.861 \\ 
        VF\tnote{2} & \textbf{0.880 }& 0.868 & 0.867 & 0.874 \\
        LE+VF\tnote{3} & 0.876 & 0.876 & \textbf{0.868} &\textbf{ 0.876 }\\ \hline
    \end{tabular}
    \begin{tablenotes}
    \footnotesize
    \item[1] LE denotes that the initial representation contains only layer's properties. 
     \item[2]VF stands for using visual features only. 
     \item[3]LE+VF means that we classify a layer from its size, type, position, and visual features.
    \end{tablenotes}
    \end{threeparttable}
\end{table}
\noindent\textbf{Feature Fusion:} Each layer has its own type, x-y coordinates, size, and visual features such as colors and shapes. Here we try to investigate which part of information plays a more important role. We initialize the initial representation vectors of nodes in the graph with three strategies. As Table \ref{ablation3} shows, we remove the CNN backbone and utilize a GNN model only to classify layers based on their types, positions, and sizes (LE). Then we conduct an experiment on detecting layers based on their visual features only (VF). Finally, we fuse these features to initialize each node's state (LE+VF).
Table \ref{ablation3} shows that the type, position, and size information can help retrieve more fragmented layers. However, the model retrieves many irrelevant layers without the help of visual features as the precision is just 81.8\%. With visual features only, the model can better guarantee the quality of retrieved layers but the recall decreases a lot by around 4\%. When combining all these features, the model has a slight increase regarding recall, accuracy, and F1-score compared with the VF method. In summary, the type, position, and size information can help to retrieve more fragmented layers while the visual features help to remove irrelevant layers.   
\label{sec:experiment}

\section{Application}

\begin{figure}[!htb]
\centering 
\includegraphics[scale=0.29]{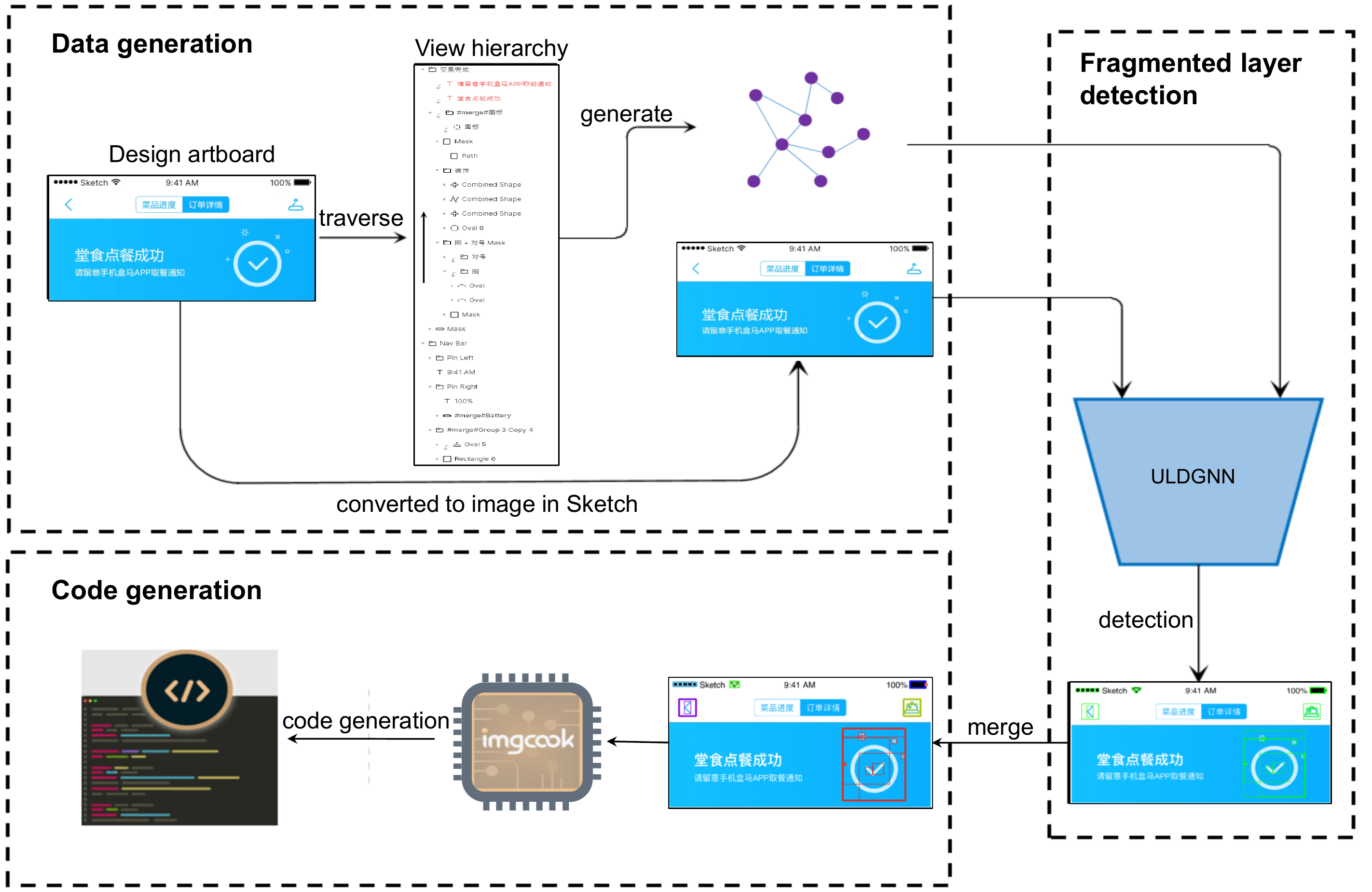} 
\caption{The whole code generation pipeline which adopts our UIDGNN model and post processing algorithm to facilitate imgcook for generating high-quality code. }
\label{application} 
\end{figure}

In this section, we utilize our proposed fragmented UI layer detector followed by a post-processing algorithm to address the issue of merging associative layers. We design a rule-based algorithm to judge whether two layers belong to the same pattern in the UI layout. Fig.\ref{application} shows how our approach can be utilized as a useful tool to correct issues in design drafts before intelligent code generation with imgcook \cite{imgcook}.
\par
\subsection{Post Processing Algorithm}
In Section 3.2, a tree in which nodes represent layers is constructed for the UI layout according to the inclusion relationship between layers. We use the tree to facilitate the layer merging process. Based on some prior knowledge, we conclude some rules: 1) layers that are close enough to each other should be merged together; 2) if a non-leaf node in the tree is classified into a positive sample, it is merged with its children with large probability. The reasons are as follows: 1) the distance is small between layers in UI icons, such as a Wifi icon; 2) designers may stack layers inside an empty layer to form a pattern as shown in Fig.\ref{mergingresult}; 3) layers in the background patterns are contained in a large layer. Therefore, our post-processing algorithm is mainly based on the distance and inclusion relations between layers.\par
We construct an adjacency matrix for all detected layers, where $A_{ij}$ denotes layer i and j should be merged together. A depth-first or breadth-first search algorithm is then utilized to find all connected components of which each stands for a merging area. Specifically, we traverse all fragmented layers detected by our model and merge layers between which the distance of the centers is below a predefined threshold. These layers are ignored in the following steps. We search the tree in a top-down manner. If a non-leaf node is predicted to be merged, it should be merged with its child nodes which are also positive samples, and we set the corresponding entries in the matrix to be 1. 
\subsection{Evaluation} 
\begin{figure}[!htb]\small
\centering
\begin{tabular}{c}
\includegraphics[scale=0.57]{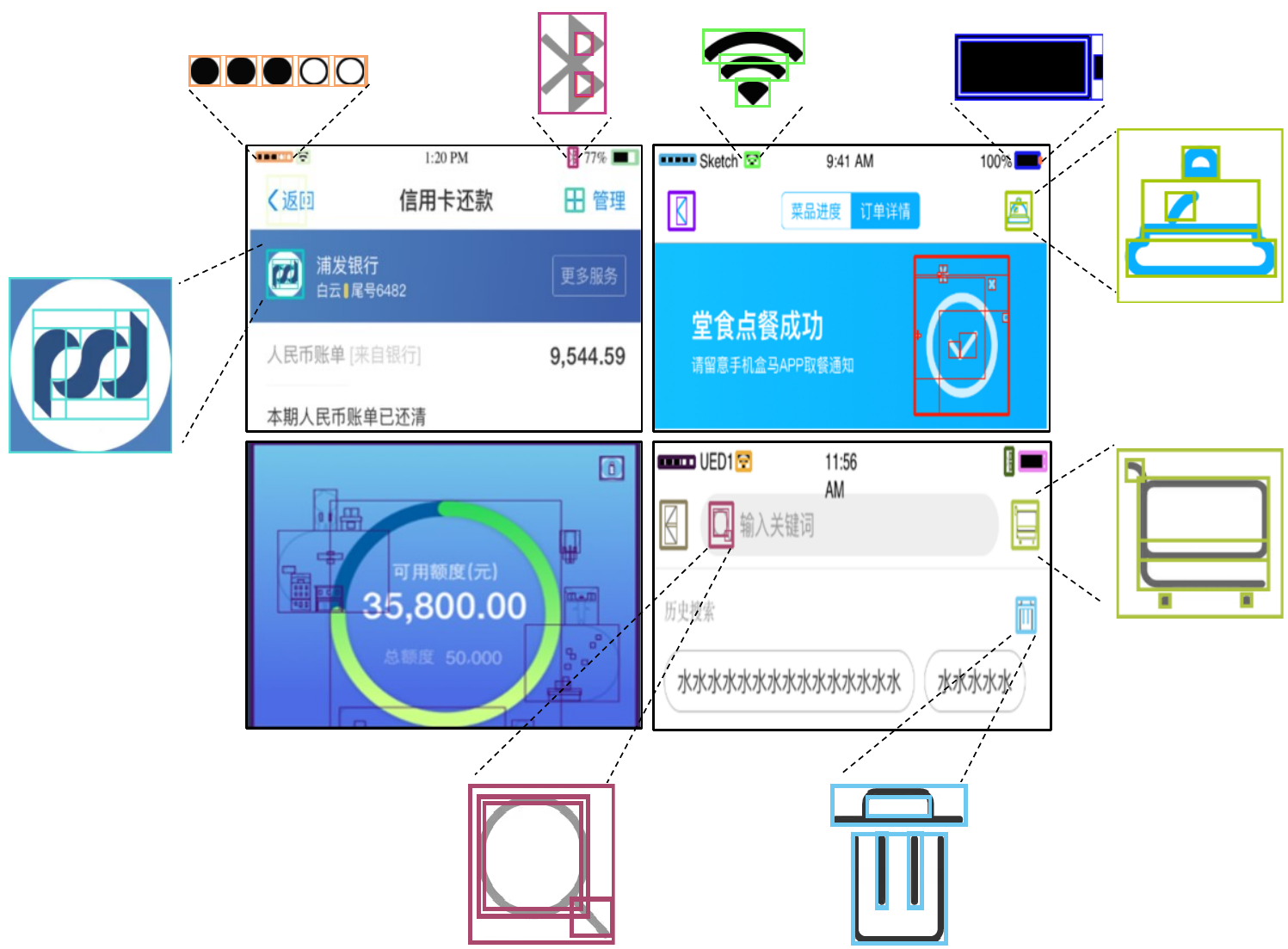}\\
{\footnotesize\sf (a) Our algorithm can merge fragmented layers in UI icons successfully. } \\[3mm]
\includegraphics[scale=0.35]{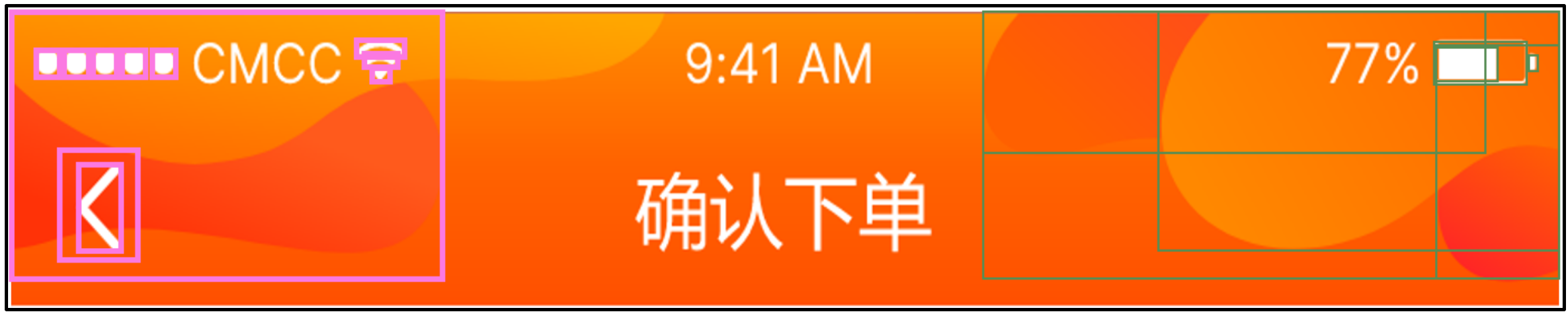}\\
{\footnotesize\sf (b) The algorithm falsely merges background layers and icon layers together. } \\
\end{tabular}
\caption{Some visualization results of our approach in a real application scenario. Layers which should be merged together are bounded with the same color. Different colors are used to distinguish different merging areas.}
\label{mergingresult}
\end{figure}

\begin{figure}[!htb]
\centering 
\includegraphics[scale=0.45]{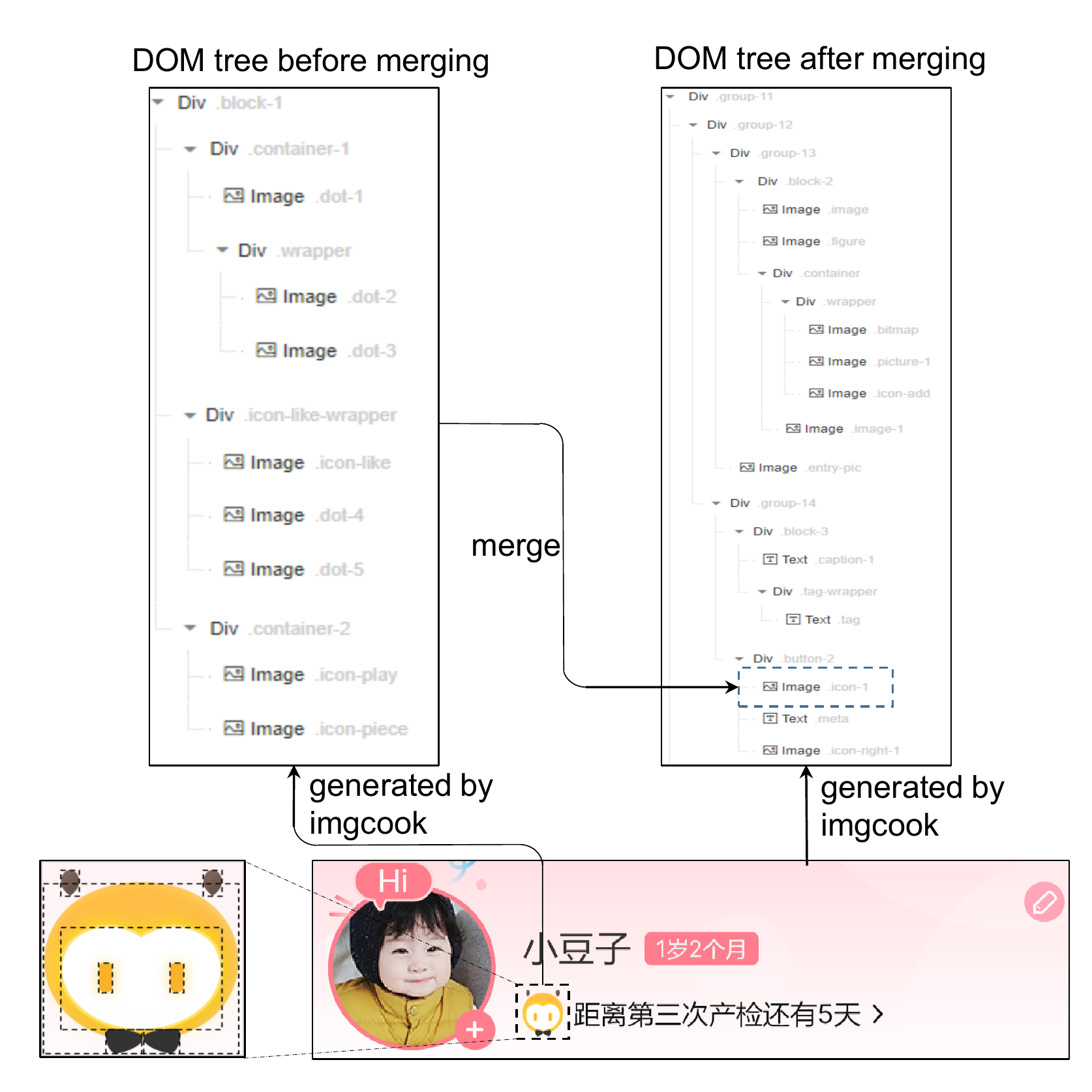} 
\caption{DOM trees generated by imgcook before and after merging fragmented layers in design drafts. }
\label{codegeneration} 
\end{figure}

We apply the proposed approach to a real application scenario where we need to merge fragmented layers into a whole part. As shown in Fig.\ref{mergingresult}a, we visualize some results of merging trivial layers in the UI layout design draft. We bound layers that should be merged together with the same color, and we use different colors to distinguish different merging areas. Our post-processing algorithm can easily merge layers in the type of UI icons and simple decorative patterns based on the rules described above. We can also discover that UI icons may be contained in an empty layer and layers in UI icons are close to each other. Designers stack many decorative layers on a large layer to form a background for the UI design draft as shown in the down-left of Fig.\ref{mergingresult}a. As our approach merges layers intelligently, which spares a lot of human effort to correct UI issues, it has a great potential in facilitating high-quality and maintainable code generation as shown in Fig.\ref{layoutmergingresult}. However, it is hard to tell the foreground and background apart as shown in Fig.\ref{mergingresult}b. The algorithm merged the "battery" icon with some background parts together as the rule-based approach ignores the semantic information. We will continue to optimize our algorithm in the future.

\label{sec:conclusion}
\section{conclusion}
We have proposed an approach to detect fragmented layers in the UI design draft based on a backbone network and a graph neural network model. We also design a rule-based post-processing algorithm to merge associative trivial layers into a whole pattern. By converting the UI layout into a graph representation, each layer can interact with each other through the GNN model and the final learned vector can better express layers' multimodal information in the UI layout. Our experiments show that the approach can easily detect and merge associative fragmented layers in icons-related UI elements but performs not very well on retrieving layers in complicated background patterns. We conduct a user study to further demonstrate the effectiveness of our approach to improve the quality of generated code.\par
In this work, we predict each layer to be positive or negative, which leaves a huge burden on the layer merging process. In the future, we can define a more concrete classification task. Specifically, a layer can be classified into one of the following types, such as a UI icon layer, a background layer, and a word font layer. 
\label{sec:conclusion}

\bibliographystyle{abbrv}
\bibliography{refs}
\end{document}